\documentclass[10pt,twocolumn,letterpaper]{article}

\usepackage{cvpr}
\usepackage{times}
\usepackage{epsfig}
\usepackage{graphicx}
\usepackage{amsmath}
\usepackage{amssymb}

\usepackage{enumerate}
\usepackage{booktabs}

\usepackage{algorithm}  
\usepackage{algpseudocode}
\usepackage{multirow}
\usepackage{mathtools}
\usepackage{mathrsfs}
\usepackage{makecell}
\usepackage{caption}
\usepackage{color,xcolor}

\newcommand*{\affaddr}[1]{#1} 
\newcommand*{\affmark}[1][*]{\textsuperscript{\textmd{#1}}}
\newcommand*{\email}[1]{\texttt{#1}}

\usepackage[breaklinks=true,bookmarks=false]{hyperref}

\cvprfinalcopy 

\def\cvprPaperID{****} 

\begin{document}

\title{Weakly-Supervised Discovery of Geometry-Aware Representation  \\ for 3D Human Pose Estimation}

\author{%
   Xipeng Chen$^\ast$\affmark[1] ~\hspace{1cm} Kwan-Yee Lin\thanks{Xipeng Chen and Kwan-Yee Lin have contributed equally and assert joint first authorship.~The work was done during the internship at SenseTime Research.} ~\affmark[2,3] \hspace{1cm} Wentao Liu\affmark[3] \hspace{1cm} Chen Qian\affmark[3]\\ Xiaogang Wang\affmark[3,4]\hspace{1cm} Liang Lin\affmark[1]\\
   \affaddr{\affmark[1]Sun Yat-Sen University}~\hspace{1cm}
   \affaddr{\affmark[2]Peking University}~\hspace{1cm}
   \affaddr{\affmark[3]SenseTime Research}\\
   \affaddr{\affmark[4]The Chinese University of Hong Kong}\\
   \small{\email{\affmark[1]chenxp37@mail2.sysu.edu.cn}}~\email{\affmark[2]linjunyi@pku.edu.cn}\\\small{\email{\affmark[3]\{liuwentao,qianchen\}@sensetime.com}}~\email{\affmark[5]xgwang@ee.cuhk.edu.hk}~\email{\affmark[4]linliang@ieee.org}\\
\vspace{-0.8cm}}

\maketitle

\begin{abstract}
Recent studies have shown remarkable advances in 3D human pose estimation from monocular images, with the help of large-scale in-door 3D datasets and sophisticated network architectures. However, the generalizability to different environments remains an elusive goal. 

In this work, we propose a geometry-aware 3D representation for the human pose to address this limitation by using multiple views in a simple auto-encoder model at the training stage and only 2D keypoint information as supervision. A view synthesis framework is proposed to learn the shared 3D representation between viewpoints with synthesizing the human pose from one viewpoint to the other one.  Instead of performing a direct transfer in the raw image-level, we propose a skeleton-based encoder-decoder mechanism to distil only pose-related representation in the latent space. A learning-based representation consistency constraint is further introduced to facilitate the robustness of latent 3D representation. Since the learnt representation encodes 3D geometry information, mapping it to 3D pose will be much easier than conventional frameworks that use an image or 2D coordinates as the input of 3D pose estimator. We demonstrate our approach on the task of 3D human pose estimation. Comprehensive experiments on three popular benchmarks show that our model can significantly improve the performance of state-of-the-art methods with simply injecting the representation as a robust 3D prior.
\end{abstract}

\section{Introduction}
3D human pose estimation refers to estimating 3D locations of body parts given an image or a video. This task is an active research topic in the computer vision community for serving as a key step for many applications, \textit{e.g.,} action recognition, human-computer interaction, and autonomous driving. Significant advances in particular datasets have been achieved in recent years due to the abundant annotations and sophisticated designed deep neural networks. However, since precise 3D annotation requires large efforts, and usually subjects to specific conditions in practice, like motions, environments, and appearances, \textit{etc.,} the bottleneck of generalizability still exists.

\begin{figure}[t]
   \begin{center}
      \includegraphics[width=1\linewidth, angle=0]{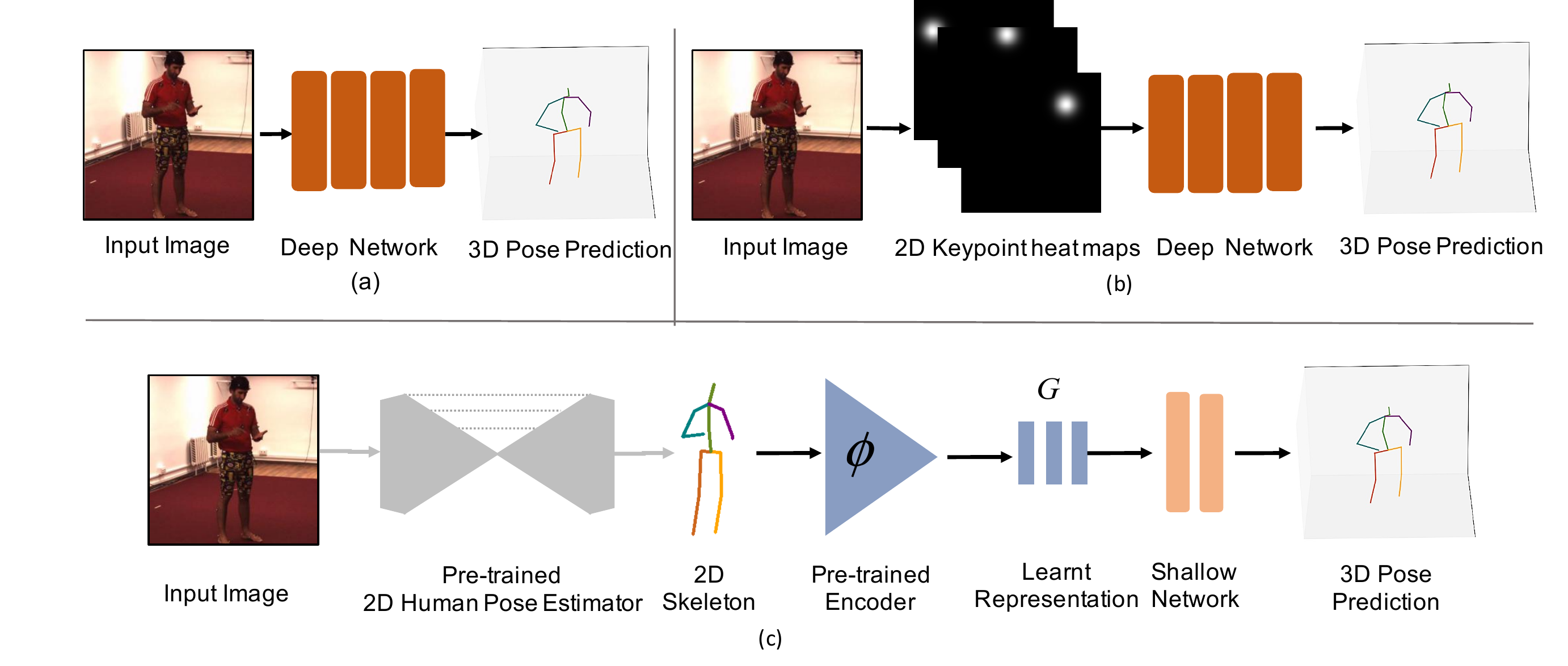}
   \end{center}
   \vskip -0.5cm
   \caption{\label{fig1}\small{Motivation. Most state-of-the-arts usually directly learn the 3D poses from monocular images (as shown in (a)), or first estimate 2D poses and then lift 2D poses to 3D poses (as shown in (b)). Both categories require sophisticated deep network architectures and abundant annotated training samples. Instead, we consider learning a geometry representation from multi-view information with only 2D annotations as supervision. The learnt representation could map to 3D pose with a shallow network and less annotated training samples, as shown in (c).}}
   \vskip -0.5cm
\end{figure}
Weakly-supervised learning provides an alternative paradigm for learning robust geometry representation without requiring extensive precise 3D annotation. Most of approaches~\cite{zhou2017towards,rhodin2018learning,popa2017deep,yang20183d,DBLP:journals/corr/abs-1901-04877} leverage knowledge transformation to learn the robustness by training 3D annotations with abundant 2D annotations in-the-wild. These methods face the difficulties of large domain shift between constrained lab environment for 3D annotations and unconstrained in-the-wild environment for 2D annotations. Some approaches try to represent body shape through multiple view images acquired by synchronized cameras with the usage of view-consistency property~\cite{rhodin2018learning}, pre-defined parametric 3D model fitting~\cite{bogo2016keep,pavlakos2018learning,MuVS:3DV:2017}, or by sequence with the usage of time-independent features~\cite{jakab2018conditional}. Nevertheless, fitting a pre-defined 3D model or exploiting limited multi-view information in a particular dataset can hardly capture all subtle poses of the human body.

The emergence of approaches for \emph{novel view synthesis}, \textit{e.g.}, \cite{flynn2016deepstereo,tulsiani2018multi},  provides an appealing and succinct solution for capturing geometry representation with multi-view information. However, despite the success of this field on many generic objects, like chairs, cars, and planes, it is \emph{non-trivial} to utilize existing frameworks to learn geometry representation for the human body, since the human body is articulated and much more deformable than rigid objects. 

The objective of this paper is to devise a simple yet effective framework that \emph{learns a 3D geometry-aware structure representation of human pose with only accessible 2D annotation as supervision}. In particular, we use an encoder-decoder to generate a novel view pose from a given view pose. The latent code of the encoder-decoder is regarded as the desired geometry representation. Instead of generating the novel view pose on \emph{image-level}~\cite{jakab2018conditional,Balakrishnan_2018_CVPR}, we propose the use of the 2D \emph{skeleton map} as a compact medium. Concretely, we first map the source and target images into 2D skeleton maps, then an encoder-decoder is trained to synthesis target skeleton from source skeleton.

Introducing the 2D skeleton as the source/target space of the encoder-decoder is beneficial for learning a robust geometry representation. Firstly, 2D skeleton could be easily obtained from an image with the usage of well-studied 2D human pose estimator~\cite{newell2016stacked,chu2017multi,DBLP:conf/aaai/LiuCL0CH18}, which is accurate and robust under diverse poses, appearances and environment conditions. This advantage could guarantee body pose and geometry information are faithfully kept. Secondly, skeleton representation avoids the variances among datasets, which could be leveraged to cover pose changes as much as possible by training existing datasets together and augment samples on continuous views. Thirdly, the representation in the latent space could be simply distilled to \emph{only} pose-related information without consideration of disentangling shape with appearance and other unessential nature of encoding geometry information.

However, the premise of obtaining a robust geometry representation under an encoder-decoder framework is the accurate generation of the target view. While, there is no theoretical assurance for generating the correct one, since the conventional view synthesis losses (\textit{e.g.,} reconstruction loss and adversarial loss) do not facilitate semantic information. To address the problem, we introduce a representation consistency loss in latent space to constrain the process without requiring any other auxiliary information. 

We summarize our contributions as follows:
\begin{enumerate}[1)]
   \item We propose a novel weakly-supervised encoder-decoder framework to learn the geometry-aware 3D representation for the human pose with multi-view data and only existing 2D annotation as supervision. To distil the representation from unessential factors, and meanwhile increase the training space, a skeleton-based view synthesis is introduced. Our approach allows the substantial 3D pose estimator to generalize well in different conditions. 
   \item To ensure the robustness of the desired representation, a representation consistency loss is introduced to constrain the learning process of latent space. In contrast to conventional weakly-supervised methods which require auxiliary information, our framework is more flexible and easier to train and implement. 
   \item A comprehensive quantitative and qualitative evaluation on public 3D human pose estimation datasets shows the significant improvements of our model applied on state-of-the-art methods, which demonstrates the effectiveness of learnt 3D geometry representation to pose estimation task.
\end{enumerate}

\begin{figure*}[t]
\begin{center}
\vskip -0.35cm
\includegraphics[width=\linewidth, angle=0]{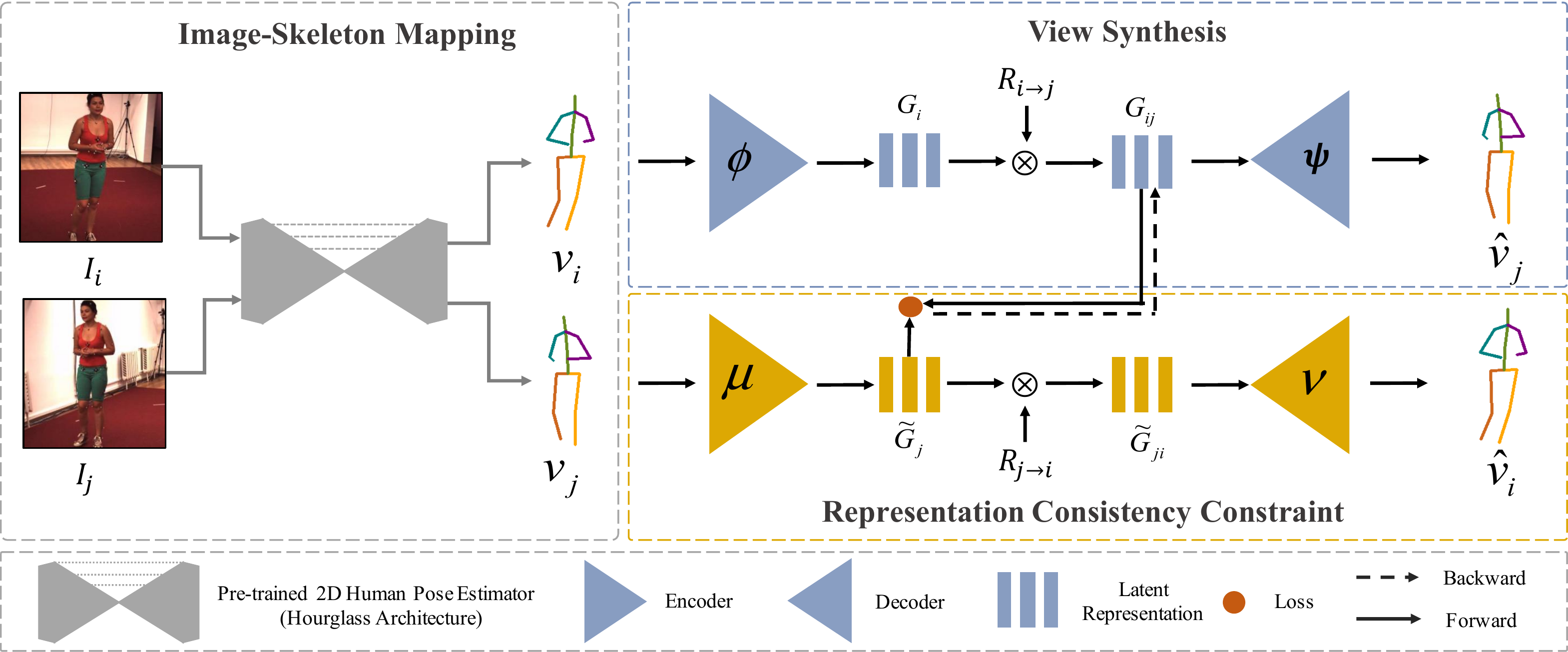}
\end{center}
\vspace{-0.5cm}
   \caption{\label{fig2}\small{ The framework of learning a geometry representation for 3D human pose in a weakly-supervised manner. There are three main components. (a)Image-skeleton mapping module is used to obtain 2D skeleton maps from raw images. (b)View synthesis module is in a position to learn the geometry representation in latent space by generating skeleton map under viewpoint $j$ from skeleton map under viewpoint $i$}. (c) Since there is no explicit constrain to facilitate the representation to be semantic, a representation consistency constrain mechanism is proposed to further refine the representation.}
\vskip -0.5cm
\end{figure*}
\section{Related Work}

\textbf{Geometry-Aware Representations.} To capture the intrinsic structure of objects, existing studies~\cite{yang2015weakly,tulsiani2018multi,jakab2018conditional,zhou2016view} typically disentangle visual content into multiple predefined factors like camera viewpoints, appearance and motion. Some works \cite{yan2016perspective,Zhang_2018_CVPR} leverage the correspondence among intra-object instance category to encode the structure representation.~\cite{Zhang_2018_CVPR} discovery landmark structure as an intermediate representation for image autoencoding with several constraints. Other approaches utilize multiple views to either directly learn the geometry representation~\cite{suwajanakorn2018discovery,Yu_2018_CVPR,Huang_2018_CVPR} with object reconstruction, or take advantage of view synthesis~\cite{poier2018learning} to learn the structure with shared latent representation between views. For example, \cite{poier2018learning} learn 3D hand pose representation by synthesizing depth maps under different views.~\cite{jakab2018conditional} conditionally generate an image of the object from another one, where the generated image differs by acquisition time or viewpoint, to encourage representation distilled to object landmarks. These methods mainly focus on structure representation of generic objects or hand/face pose. Whereas, the human body is articulated and much more deformable. How to capture the geometry representation of the human body with fewer data and simpler constraints is still an open question.

\textbf{3D Human Pose Estimation.} Most of the existing studies for 3D human pose estimation benefit from the availability of large-scale datasets and sophisticated deep-net architectures. These methods could be roughly categorized into fully-supervised and weakly-supervised manners.

A vast amount of fully-supervised 3D pose estimation methods via monocular image exist in the literature~\cite{martinez2017simple,moreno20173d,chen20173d,wang2018drpose3d}. Despite the performance these methods achieve, modeling 3D mapping from a given dataset limits their generalizability due to the constrained lab environment, limited motion and inter-dataset variation.{\footnote{Inter-dataset variation refers to bias among different datasets on viewpoints, environments, the definition of 3D key points, \textit{ etc.}}} 

Several works focus on weakly-supervised learning to increase the diversity of samples and meanwhile restrain the usage of labeled 3d annotated data. For example, synthesize training data by deforming a human template model with known 3D ground truth~\cite{varol2017learning}, or generating various foreground/background~\cite{mehta2017monocular}.~\cite{zhou2017towards} proposes to transform knowledge from 2D pose to 3d pose estimation network with re-projection constraint to 2D results. A converse strategy is employed in \cite{yang20183d} to distil 3D pose structure to unconstrained domain under an adversarial learning framework.~\cite{pavlakos2018learning} proposes to learn the parameters of the statistical model SMPL \cite{loper2015smpl} to obtain 3D mesh from image with an end-to-end network, and regresses 3d coordinates from the mesh. Other approaches~\cite{rhodin2018learning,DBLP:journals/corr/abs-1712-05765} exploit views consistency with the usage of multiple viewpoints of the same person. Nevertheless, these methods still rely on a large quantity of 3D training samples or auxiliary annotations, like silhouettes \cite{du2016marker} and depth~\cite{DBLP:journals/corr/abs-1712-05765} to initialize or constrain the models.

In contrast to above approaches, our framework aims at discovering a robust geometry-aware 3D representation of human pose in latent space, with only 2D annotation in hand. This allows us to train the subsequent monocular 3D pose estimation network with much less labeled 3D data. Recently, a concurrent work is published in the community with similar spirits. In contrast to \cite{rhodin2018unsupervised} that can only handle one particular dataset due to the dependency of appearance and inter-frame information during the training process, our framework tries to break the gap of inter-dataset variation, which permits more practical usages. Moreover, our framework is complementary to previous 3D pose estimation works, and can use current approaches as the baseline with the injection of learnt representation as a 3D structure prior.
\vspace{-0.3cm}
\section{Weakly-Supervised Geometry Representation}
\vspace{-0.2cm}
Recall that our goal is to learn a geometry-aware 3D representation $\mathcal{G}$ for the human pose, which is expected to be \emph{robust} to diverse pose changes and can be learnt with \emph{less effort} than conventional weakly-supervised methods. Toward this end, we propose to discover the geometry relation between paired images$(I_t^i, I_t^j)$, which are acquired from \emph{synchronized} and calibrated cameras, with the only \emph{existing} 2D coordinate annotation used for supervision, where $i$ and $j$ denote different viewpoints, $t$ denotes acquiring time. The proposed approach is depicted in Figure~\ref{fig2}. The framework includes three components: an image-skeleton mapping component, a skeleton-based view synthesis component, and a representation consistency constraint component. The desired representation is encoded in the \emph{bottleneck} of the encoder-decoder on the view synthesis component. In the inference phase, the learnt representation will be obtained by forwarding a \emph{single} image through the first two components, as illustrated in Figure~\ref{fig1}(c). We will detail each component in the remainder of this section.

\subsection{Image-skeleton mapping}
It is habitual to directly feed forward the raw image to the network to learn geometry representation~\cite{jakab2018conditional,tulsiani2018multi}. However, under the setting of multiple-view with encoder-decoder framework, we demonstrate that utilizing only 2D skeleton information is sufficient and better than raw images to learn the representation, as shown in the Sec~\ref{exp}. Consequently, given a pair of raw images $(I_t^i, I_t^j)$ with the size of $W \times H$ under different viewpoints of camera $i$ and camera $j$ respectively, a pre-trained 2D human pose estimator{\footnote{We follow previous works~\cite{zhou2017towards,moreno20173d,martinez2017simple} to train the 2D estimator on MPII dataset.}} is firstly applied to obtain two stacks of $K$ key point heatmaps $C_t^i$, and $C_t^j$. Then, the corresponding 2D skeleton maps, regarded as a person tree-structured kinematic graph, are constructed from the heatmaps with $8$ pixels width. Consequently, we are given the binary skeleton maps pair $(S_t^i,S_t^j)$, where $S_t^{(\cdot)} \in \{0,1\}^{(K-1)\times W\times H}$.

Intuitively, we could sample $(i,j)$ randomly from existing cameras. However, such a sampling strategy will lead to two problems in practice. Firstly, the finite samples limit the diversity of the training set. Secondly, the nonuniform distribution{\footnote{For example, in Human3.6M dataset~\cite{h36m_pami}, four cameras are approximately located at four corners of a rectangle.}} of viewpoints will increase the difficulty of network learning. To solve the above problems, it is straightforward to utilize virtual cameras-based data augmentation. While, conventional methods can only achieve in-plane rotations due to image-level inputs~\cite{jakab2018conditional,rhodin2018unsupervised}. 
Instead, we draw on virtual cameras applied in~\cite{fang2018learning} to increase training pairs on a torus{\footnote{Please refer to the supplemental materials for detail operation.}}. Different from~\cite{fang2018learning} that generate new 2D coordinates-3D coordinates pairs, we randomly sample 2D skeleton pairs. Thus, we could obtain infinite training pairs and calculate their relative rotation matrix in theory. This augmentation strategy facilitates our model to be robust to different camera configurations.

\subsection{Geometry representation via view synthesis}
Assume that we are given a training set $\mathcal{T}=\{(S_t^i,S_t^j,R_{i\rightarrow j})\}_{t=1}^{N_T}$ containing pairs of two views of projection of same 3D skeleton $(S_t^i,S_t^j)$  and relative rotation matrix $R_{i\rightarrow j}$ from coordinate system of camera $i$ to $j$, after image-skeleton mapping step. We now turn to discover the geometry representation $\mathcal{G}$. A straightforward way for learning representation in unsupervised/weakly-supervised manner is to utilize autoencoding mechanism reconstructing input image. Then, the latent codes of the auto-encoder could be regarded as the features that encode compact information of the input~\cite{Zhang_2018_CVPR,DBLP:journals/corr/KingmaW13}. While, such a representation neither contains geometry structure information nor provides more useful information for 3D pose estimation than 2D coordinates, as demonstrated in Figure~\ref{fig5}.

The proposed ~`skeleton-based view synthesis' step draws an idea from novel view synthesis methods, which usually rely on the encoder-decoder framework to generate image under a new viewpoint of the same object, given an image under the known viewpoint as input.
Without the loss of generality, the input images are regarded as the \emph{source domain}, and the generated ones are regarded as the \emph{target domain}. We tailor the process to our problem as follows. 

Let $\mathcal{S}^i=\{S_t^i\}_{i=1}^V$ be the source domain, where $V$ denotes the amount of viewpoints, and $\mathcal{S}^j=\{S_t^j\}_{j=1}^V$ be the target domain with $j\neq i$. We are interested in learning an encoder $\phi:\mathcal{S}^i\rightarrow \mathcal{G}$ that capture the geometry structure of the human pose. The encoder maps a source skeleton $S_t^i \in \mathcal{S}^i$ into a latent space $G_{i}\in \mathcal{G}$. In order to learn $\mathcal{G}$, the property of the \emph{shared} representation between the source and target domains should be satisfied. Thus, under the control of relative rotation matrix $R_{i\rightarrow j}$, $G_{i}$ should be decoded back to the target view with a decoder $\psi:\mathcal{R}_{i\rightarrow j} \times \mathcal{G} \rightarrow \mathcal{S}^j$. Besides, if $\mathcal{G}$ is close to the manifold of 3D pose coordinates, the learning process of subsequent monocular 3D pose estimation will be simplified and less labeled 3D data will be needed. So far, it is difficult to demonstrate whether the learnt $G_{i}$ satisfy the assumption, since the framework doesn't contain any explicit constraint to $G_{i}$. To this end, the dimensional space of $\mathcal{G}$ should be constrained at first. We formulate $G_{i}$ as the set of $m$ discrete points on a $3\eta$-dimensional feature space with the form of a $3\eta$-dimensional and $M$-length feature vector in practice, \textit{i.e.,} $G=[g_1,g_2,\cdots,g_M]^\top$ with $g_m=(x_m,y_m,z_m)$. We adopt $L2$ reconstruction loss to the learning process:
\begin{equation}\label{eq1}
L_{\ell_2}(\phi \cdot \psi,\theta) =\frac{1}{N^T} \sum\|\psi(R_{i\rightarrow j}\times \phi(S_t^i))- S_t^j\|^2.
\end{equation}

While the combination of reconstruction loss, adversarial loss and perceptual loss are widely used in synthesis tasks~\cite{Balakrishnan_2018_CVPR,DBLP:journals/corr/abs-1807-11079,wayne2018lab}, the rest two losses will introduce artificial noise to our framework. Since skeleton maps only contain low-frequency information when regarded as the images.
\begin{figure}[t]
   \begin{center}
      
      \includegraphics[width=1\linewidth, angle=0]{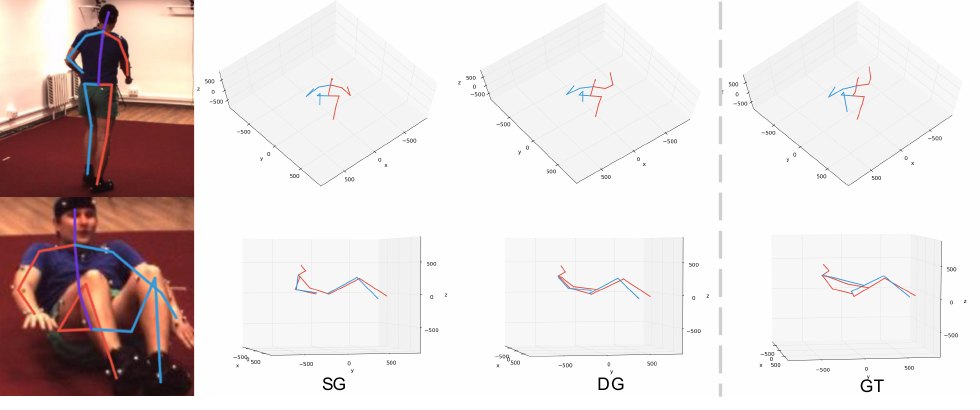}
   \end{center}
   \vspace {-0.4cm}
   \caption{\label{fig3} \small{An illustration of the effectiveness of representation consistency constraint. Compared with only applying the~`image-skeleton mapping+view synthesis'(SG), the representation consistency constraint(DG) is able to refine the implausible poses, which is more similar to the ground-truth poses(GT)(zoom in for more details)}. }
   \vskip -0.5cm
\end{figure}


\subsection{Representation consistency constraint}
As shown in Figure~\ref{fig3}, only applying ~`image-skeleton mapping+view synthesis' components may lead to unrealistic generation on target pose when there are large self-occlusions on source view, which will lead the learnt representation $\mathcal{G}$ misleading the regression of subsequent 3D pose estimation task. Since there is no explicit constraint on latent space to facilitate $\mathcal{G}$ to be semantic. To this end, we propose a representation consistency constraint to the framework. We assume there exists an inverse mapping (one-to-one) between source domain and target domain, on the condition of the known relative rotation matrix. Then, we could find an encoder $\mathcal{\mu}:\mathcal{S}^j\rightarrow \mathcal{G}$  maps target skeleton $S_t^j$ to the latent space $\tilde{G_j}\in{\mathcal{G}}$, and a decoder $\nu:R_{j\rightarrow i} \times G \rightarrow S^i$ maps the representation $\tilde{G_j}$
back to source skeleton $S_t^i$ on the condition of ${R}_{j\rightarrow i}$. Thus, for paired data $(S_t^i,S_t^j)$, $G_i$ and $\tilde{G_j}$ should be the same shared representation on $\mathcal{G}$ with different rotation-related coefficients. We add this relationship, namely representation consistency, to the network explicitly with the formulation as:
\begin{equation}\label{eq2}
l_{rc} =\sum_{m=1}^M\|f\times G_i-\tilde{G_j}\|^2,
\end{equation}

where $f$ denotes the rotation-related transformation that map $G_i$ to $\tilde{G_j}$. This loss function is well-defined when $f$ is known. To release the constraint, we simply assume $f= R_{i\rightarrow j}$. In practice, we implement the representation consistency constraint by designing a bidirectional encoder-decoder framework, which hinges on two encoder-decoder networks with same architecture, \textit{i.e.,} $generator(\mathcal{\phi},\mathcal{\psi})$ and $generator(\mathcal{\mu},\mathcal{\nu})$, to perform view synthesis in the two directions simultaneously. Specifically, let $G_{ij}$ be the rotated $G_i$ on $generator(\mathcal{\phi},\mathcal{\psi})$-branch, we enforce normalized $G_{ij}$ to be close to normalized $\tilde{G_j}$ with modified Eqn~\ref{eq2}:
\begin{equation}
L_{rc}(\mathcal{\phi},\mathcal{\mu},\zeta) = \sum_{m=1}^M\|g_{ij_m}- g_{j_m}\|_2^2 .
\end{equation}
The general idea behind the formula is that if the mapping could be perfectly modeling, the latent codes $G_i$ and $\tilde{G_j}$ would be
the same geometry representation under world coordinate system mapping to different camera coordinate systems. In other words, the consistency constraint enforces the learnt latent codes containing explicit~\textit{physical meanings}. Thus, features of implausible poses could be distilled. With more robust representation, subsequent pose estimation results will be improved. 

\begin{figure}[t]
   \begin{center}
      
      \includegraphics[width=1\linewidth, angle=0]{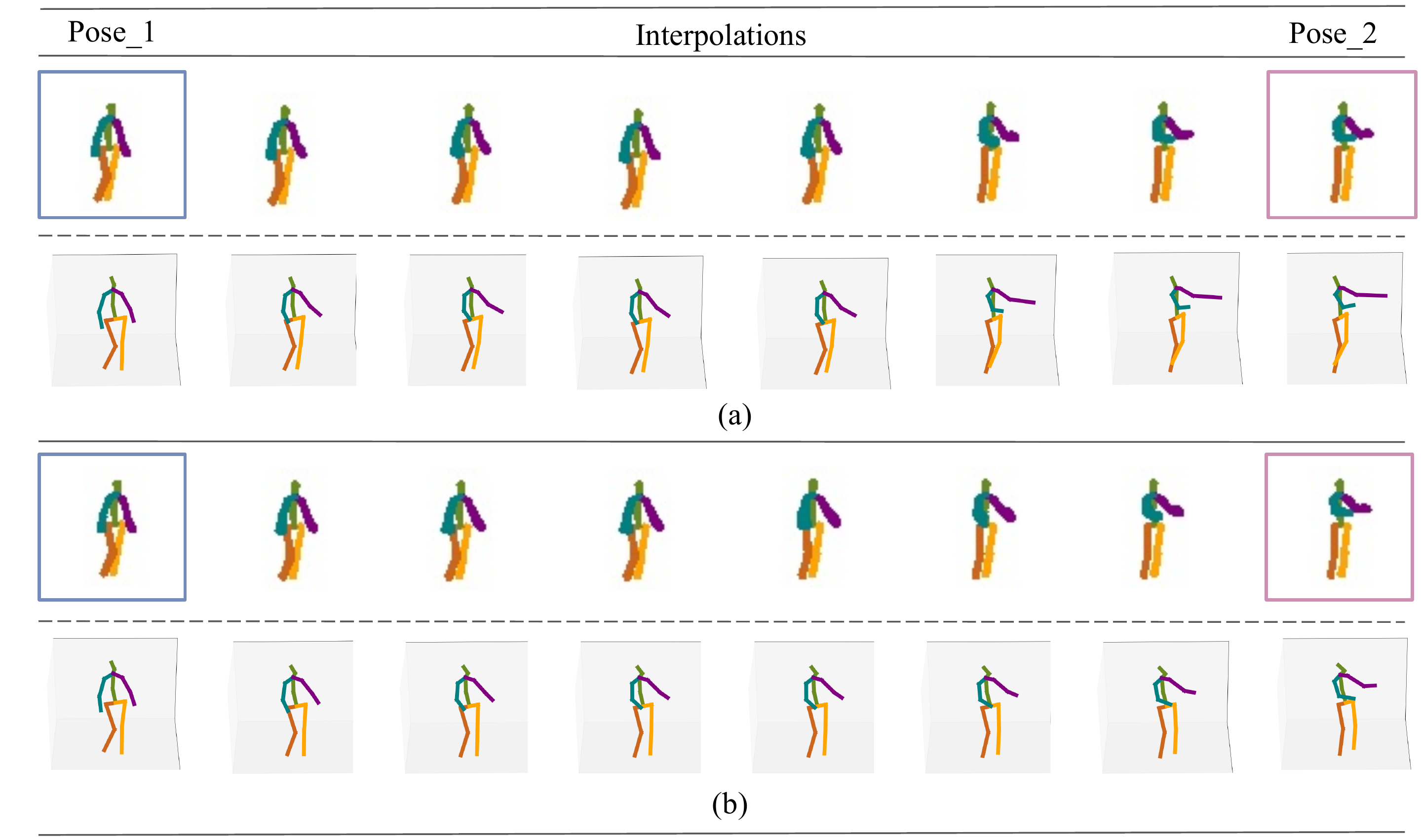}
   \end{center}
   \vspace {-0.6cm}
   \caption{\label{point} \small{Illustration of point cloud (the learnt latent representation) interpolation. Pose$\_1$ and Pose$\_2$ are two randomly sampled poses under same camera viewpoint. (a) and (b) show the interpolation results of the latent codes learned without/with representation constraint, respectively. There are two main differences between (a) and (b). First, from first rows in (a) and (b), (b) shows more smooth interpolation results (for example, the change of arms from the fifth column to the sixth column), than the ones in (a). Second, the lower part of the body should gradually stand upright and spraddle from left to right for both 2D skeleton and 3D pose. However, it is inconsistent between the 2D skeleton and 3D pose in (a). Instead, the results in (b) are consistent.} }
   \vspace{-0.4cm}
\end{figure}

Besides, since the latent codes are formulated as the set of $m$ discrete points on a $3\eta$-dimensional feature space, they could be regarded as $3D$ point clouds. In Figure~\ref{point}, we show both point clouds interpolations with/without proposed representation constraint to illustrate the claim qualitatively. As can be seen from the figure, the linear interpolation results of the one with representation constraint show more reasonable coverage of the manifold, and better consistency between decoded $2D$ skeleton on the target domain and regressed $3D$ pose. This phenomenon demonstrates the learnt latent codes have extracted better $3D$ geometry representations of the human shape with the help of representation constraint.

We train our bidirectional model in an end-to-end manner, minimizing the following total loss:
\begin{equation}
\mathcal{L} = L_{\ell_2}(\phi \cdot \psi,\theta) + L_{\ell_2}(\mu \cdot \nu,\zeta) +L_{rc}(\mathcal{\phi},\mathcal{\mu}),
\end{equation}
where $\theta$ and $\zeta$ denotes the parameters of two encode-decoder networks, respectively.

\subsection{3D human pose estimation by learnt representation}
Recall that our ultimate goal is to inference 3D human pose in the form of $\mathbf{b}=\{(x^p,y^p,z^p)\}_{p=1}^P$ from a monocular image $I$, where $P$ denotes the amount of body joint locations, and $\mathbf{b}\in \mathcal{B}$. In this section, we discuss how to find a function $\mathcal{F}:\mathcal{I}\rightarrow \mathcal{B}$ to learn the pose regression. Above components first lift the raw image to a 2D skeleton representation, then the 2D skeleton is lifted to $G$, which is a $3D$ geometry representation for human body. Thus, we could split function $\mathcal{F}$ into three sub-functions: $\mathcal{F}_{2D}$, $\mathcal{F}_G$ and $\mathcal{F}_{regression}$, with:
\begin{equation}
\mathcal{F}(I) =\mathcal{F}_{regression}(\mathcal{F}_{G}(\mathcal{F}_{2D}(I)))=  \mathcal{F}_{regression}(G),
\end{equation}
where $\mathcal{F}_{2D}$ denotes the first component, and $\mathcal{F}_{G}$ denotes the second component. Since $G\in \mathbb{R}^{3\times M}$ and $\mathbf{b} \in \mathbb{R}^{3\times P}$, $\mathcal{F}_{regression}(\cdot)$ could be a linear function to decode $G$ to $\mathbf{b}$. In practice, we implement the regression part by simply constructing a two-layers fully-connected neural network. Specifically, we firstly feed forward the raw image to the \emph{fixed} components ~`image-skeleton mapping+$\mathcal{\phi}$' to obtain $G$, then $G$ is regarded as the input to $\mathcal{F}_{regression}(\cdot)$ to regress the final coordinates. Only leveraging a small set of labeled samples to train the regression part could lead to satisfied accuracy, as demonstrated in Sec~\ref{exp}.

\begin{table*}[t]
   \setlength{\abovecaptionskip}{0.1cm}
   \setlength{\belowcaptionskip}{-0.5cm}
   \centering
   \small
   \setlength{\intextsep}{0cm}
   \setlength{\textfloatsep}{0pt}
   \resizebox{
      1.\textwidth}{!}{
      \begin{tabular}{@{}lcccccccccccccccc@{}}
         \specialrule{1pt}{1pt}{2pt}
         Protocol \#1 & Direction & Discuss & Eat & Greet & Phone & Photo & Pose & Purchase & Sit & SitDown & Smoke & Wait & WalkDog & Walk & WalkT. & Avg.\\
         \specialrule{0.5pt}{0pt}{1pt}

         Zhou \etal (ICCV'17)~\cite{zhou2017towards}      & 54.8          & 60.7          & 58.2          & 71.4          & 62.0 & 65.5 & 53.8          & 55.6          & 75.2          & 111.6         & 64.1          & 66.0          & 51.4          & 63.2          & 55.3          & 64.9          \\
         Martinez \etal (ICCV'17)~\cite{martinez2017simple}  & 51.8          & 56.2          & 58.1          & 59.0          & 69.5        & 78.4          & 55.2          & 58.1          & 74.0          & 94.6          & 62.3          & 59.1          & 65.1          & 49.5          & 52.4          & 62.9          \\
         Fang \etal (AAAI'18)~\cite{fang2018learning}      & 50.1          & 54.3          & 57.0          & 57.1          & 66.6        & 73.3          & 53.4          & 55.7          & 72.8          & 88.6          & 60.3          & 57.7          & 62.7          & 47.5          & 50.6          & 60.4          \\
         Sun \etal (ICCV'17)~\cite{sun2017compositional}      & 52.8      & 54.8    & 54.2   & 54.3  & 61.8  & 67.2  & 53.1 & 53.6     & 71.7    & 86.7        & 61.5  & 53.4 & 61.6   & 47.1 & 53.4   & 59.1 \\
         
         Yang \etal (CVPR'18)~\cite{yang20183d}         & 51.5           & 58.9       & 50.4          & 57.0          & 62.1           & 65.4       & 49.8          & 52.7          & 69.2           & 85.2       & 57.4          & 58.4          & 43.6           & 60.1       & 47.7          & 58.6      \\

         Pavlakos \etal (CVPR'18)~\cite{Pavlakos_2018_CVPR}  & 48.5           & 54.4       & 54.4          & 52.0         & 59.4           & 65.3       & 49.9         & 52.9          & 65.8           & 71.1      & 56.6        & 52.9         & 60.9           & 44.7       & 47.8          & 56.2      \\

         Wang \etal (IJCAI'18)~\cite{wang2018drpose3d}        & 49.2           & 55.5       & 53.6          & 53.4          & 63.8           & 67.7       &  50.2          & 51.9          & 70.3           & 81.5       &  57.7          & 51.5          & 58.6           & 44.6       &  47.2          & 57.8      \\
         
         Trumble \etal (ECCV'18)~\cite{DBLP:conf/eccv/TrumbleGHC18}      & 41.7           & 43.2        & 52.9          & 70.0          & 64.9           & 83.0        & 57.3          & 63.5          & 61.0           & 95.0        & 70.0          & 62.3          & 66.2           & 53.7        & 52.4          & 62.5      \\
         
         Park \etal (BMVC'18)~\cite{DBLP:conf/bmvc/ParkK18}         & 49.4           & 54.3        & 51.6          & 55.0          & 61.0           & 73.3        & 53.7          & 50.0          & 68.5           & 88.7        & 58.6          & 56.8          & 57.8           & 46.2        & 48.6          & 58.6         \\

         Sun \etal (ECCV'18)~\cite{DBLP:conf/eccv/SunXWLW18}        & 46.5         & 48.1          & 49.9       & 51.1          & 47.3          & 43.2           & 45.9        & 57.0        & 77.6          & 47.9           & 54.9        & 46.9          & 37.1          & 49.8           & 41.2       & 49.8             \\
         
         \specialrule{0.5pt}{0pt}{2pt}
         Ours + Regression\#1 (2 fc layers)         & 63.9          & 73.7           & 70.9        & 76.1          & 82.6          & 69.5          & 75.1       & 96.1        & 120.6        & 75.4          & 96.8        & 78.7          & 69.1          & 83.5           & 72.2       & 80.2             \\

         Ours + Regression\#2 (~\cite{martinez2017simple})          & 45.9          & 53.5           & 50.1        & 53.2          & 61.5          & 72.8           & 50.7        & 49.4          & 68.4          & 82.1           & 58.6        & 53.9          & 57.6          & 41.1           & 46.0        & 56.9             \\
         
         Ours + Regression\#3 (~\cite{DBLP:conf/eccv/SunXWLW18})          & 41.1          & 44.2           & 44.9        & 45.9          & 46.5          & 39.3           & 41.6        & 54.8          & 73.2          & 46.2           & 48.7        & 42.1          & 35.8          & 46.6           & 38.5        & \textbf{46.3}           \\

         \specialrule{1pt}{1pt}{2pt}
         Protocol \#2                & Direction & Discuss & Eat & Greet & Phone & Photo & Pose & Purchase & Sit & SitDown & Smoke & Wait & WalkDog & Walk & WalkT. & Avg. \\
         \specialrule{0.5pt}{0pt}{1pt}

         Moreno-Noguer (CVPR'17)~\cite{moreno20173d}   & 66.1      & 61.7    & 84.5   & 73.7  & 65.2  & 67.2  & 60.9 & 67.3     & 103.5   & 74.6        & 92.6  & 69.6 & 71.5   & 78.0 & 73.2   & 74.0 \\
         
         Zhou et al. (Arxiv'17)~\cite{DBLP:journals/corr/ZhouZPLDD17}    & 47.9      & 48.8    & 52.7   & 55.0  & 56.8  & 65.5  & 49.0 & 45.5     & 60.8    & 81.1        & 53.7  & 51.6 & 54.8   & 50.4 & 55.9   & 55.3 \\
         
         Sun et al. (ICCV'17)~\cite{sun2017compositional}      & 42.1      & 44.3    & 45.0   & 45.4  & 51.5  & 53.0  & 43.2 & 41.3     & 59.3    & 73.3        & 51.0  & 44.0 & 48.0   & 38.3 & 44.8   & 48.3 \\
         
         Martinez et al. (ICCV'17)~\cite{martinez2017simple} & 39.5      & 43.2    & 46.4   & 47.0  & 51.0  & 56.0  & 41.4 & 40.6     & 56.5    & 69.4        & 49.2  & 45.0 & 49.5   & 38.0 & 43.1   & 47.7 \\
         
         Fang et al. (AAAI'18)~\cite{fang2018learning}      & 38.2      & 41.7    & 43.7   & 44.9  & 48.5  & 55.3  & 40.2 & 38.2     & 54.5    & 64.4        & 47.2  & 44.3 & 47.3   & 36.7 & 41.7   & 45.7 \\
         
         Sun \etal (ECCV'18)~\cite{DBLP:conf/eccv/SunXWLW18}        & 40.9         & 41.4           & 45.0        & 45.2          & 42.1          & 37.6           & 41.1        & 52.0        & 71.4          & 42.5           & 47.4        & 41.6          & 32.0         & 42.6           & 36.9        & 44.1             \\

         Yang et al. (CVPR'18)~\cite{yang20183d}         & 26.9           & 30.9        & 36.3         & 39.9           & 43.9           & 47.4        & 28.8         & 29.4           & 36.9           & 58.4        & 41.5         & 30.5           & 29.5           & 42.5        & 32.2         & \textbf{37.7}           \\
         
         \specialrule{0.5pt}{0pt}{2pt}
         Ours + Regression\#1 (2 fc layers)         & 47.0          & 51.8           & 53.3       & 55.3          & 59.7         & 48.4           & 51.7        & 72.1        & 90.6          & 56.6           & 65.4        & 55.1          & 50.2         & 59.4           & 53.9        & 58.2             \\

         Ours  + Regression\#2 (\cite{martinez2017simple})       & 36.5           & 41.0           & 40.9        & 43.9         & 45.6           & 53.8           & 38.5        & 37.3         & 53.0           & 65.2           & 44.6        & 40.9         & 44.3           & 32.0           & 38.4        & 44.1                         \\

         Ours + Regression\#3 (\cite{DBLP:conf/eccv/SunXWLW18})         & 36.9          & 39.3          & 40.5        & 41.2         & 42.0          & 34.9           & 38.0       & 51.2        & 67.5          & 42.1           & 42.5        & 37.5          & 30.6          & 40.2           & 34.2        & 41.6             \\

         \specialrule{1pt}{1pt}{2pt}
         
         \specialrule{1pt}{1pt}{2pt}
         Protocol \#3                & Direction     & Discuss       & Eat           & Greet         & Phone         & Photo         & Pose          & Purchase      & Sit           & SitDown        & Smoke         & Wait          & WalkDog       & Walk        & WalkT.        & Avg.        \\
         \specialrule{0.5pt}{0pt}{1pt}
         Pavlakos et al. (CVPR'17)~\cite{pavlakos2017coarse} & 79.2          & 85.2          & 78.3          & 89.9          & 86.3          & 87.9          & 75.8          & 81.8          & 106.4         & 137.6          & 86.2          & 92.3          & 72.9          & 82.3        & 77.5          & 88.6        \\
         Martinez et al. (ICCV'17)~\cite{martinez2017simple} & 65.7          & 68.8          & 92.6          & 79.9          & 84.5          & 100.4         & 72.3          & 88.2          & 109.5         & 130.8          & 76.9          & 81.4          & 85.5          & 69.1        & 68.2          & 84.9        \\
         Zhou et al. (ICCV'17)~\cite{zhou2017towards}     & 61.4          & 70.7          & 62.2          & 76.9          & 71.0          & 81.2          & 67.3          & 71.6          & 96.7          & 126.1          & 68.1          & 76.7          & 63.3          & 72.1        & 68.9          & 75.6        \\
         Fang et al. (AAAI'18)~\cite{fang2018learning}      & 57.5          & 57.8          & 81.6          & 68.8          & 75.1          & 85.8          & 61.6          & 70.4          & 95.8          & 106.9 & 68.5          & 70.4          & 73.89         & 58.5        & 59.6          & 72.8        \\

         Sun \etal (ECCV'18)~\cite{DBLP:conf/eccv/SunXWLW18}        & 52.4         & 50.5           & 45.0        & 57.8          & 49.8          & 50.3          & 46.1       & 57.1        & 96.3          & 47.4           & 56.4        & 52.1          & 45.7          & 53.7           & 48.7        & 53.6             \\

         \specialrule{0.5pt}{0pt}{1pt}
         Ours + Regression\#1 (2 fc layers)          & 70.8         & 78.3           & 84.9        & 89.2          & 89.2          & 78.0           & 85.6        & 116.3        & 142.7          & 87.0          & 114.2       & 88.1          & 81.5          & 92.9          & 80.3        & 91.4            \\
         
         Ours + Regression\#2 (\cite{martinez2017simple})        & 60.4          & 63.6           & 77.2        & 69.5          & 64.8          & 96.1           & 64.1        & 75.0          & 87.6          & 111.1           & 66.6        & 67.7          & 70.0          & 54.8           & 57.6        & \textbf71.8           \\ 

         Ours + Regression\#3 (\cite{DBLP:conf/eccv/SunXWLW18})          & 45.9          & 48.0           & 48.6        & 50.8          & 48.9          & 45.1           & 46.1        & 57.4        & 77.3          & 49.4           & 54.2        & 47.2          & 39.9          & 49.9          & 42.9        & \textbf{50.3}             \\  
         
         \specialrule{1pt}{1pt}{2pt}

      \end{tabular}
   }
   \vspace {-0.1cm}
   \caption{\label{tb1}\small{Quantitative comparisons of Mean Per Joint Position Error ($mm$) between the estimated pose and the ground-truth on Human3.6M under Protocol \#1,\#2 \#3. The best score is marked in bold.}}
   \label{human3.6M}
   \vspace {-0.2cm}        
\end{table*}

\section{Experiments}\label{exp}

\textbf{Datasets.} We evaluate our approach both quantitatively and qualitatively on popular human pose estimation benchmarks: Human3.6M~\cite{h36m_pami}, MPI-INF-3DHP~\cite{mehta2017monocular}, and MPII Human Pose~\cite{andriluka14cvpr}. Human3.6M is the largest dataset for 3D human pose estimation, which consists of $3.6$ million poses and corresponding video frames featuring $11$ actors performing $15$ daily activities from $4$ camera views. MPI-INF-3DHP is a recently proposed 3D benchmark consists of both constrained indoor and complex outdoor scenes. MPII Human Pose dataset is a challenging benchmark for estimating in-the-wild 2D human pose. Following previous methods~\cite{yang20183d,fang2018learning,pavlakos2017coarse,martinez2017simple}, we adopt this dataset for evaluating the cross-domain generalization qualitatively.

\textbf{Evaluation Protocols.} For Human3.6M dataset, we follow the standard protocol, \textit{i.e.,\textit{Protocol\#1}}, to use all $4$ camera views in subjects $1$, $5$, $6$, $7$ and $8$ for training, and same all $4$ camera views in $9$ and $11$ for testing. In some works, the predictions are further aligned with the ground-truth via a rigid transformation~\cite{yang20183d,fang2018learning}, which is referred as \textit{Protocol\#2}. To further validate the robustness of different models to new subjects and views, we follow~\cite{fang2018learning} to use subjects $1$, $5$, $6$, $7$ and $8$ in $3$ camera views for training, while $9$ and $11$ in the other camera view for testing. This protocol is referred as \textit{Protocol\#3}. The evaluation metric is the Mean Per Joint Position Error (MPJPE), measured in millimeters.

\textbf{Implementation Details.}
For ~`image-skeleton mapping' module, we adopt a state-of-the-art 2D pose estimator~\cite{newell2016stacked} to perform 2D pose detection. We adopt the network architecture on the U-Net as the backbone of our $generator(\cdot,\cdot)$. The skip connections are removed to ensure all information can be encoded into the latent codes. For model acceleration, we also halve the feature channels and modify the input and output to 15-channel $64 \times 64$. The regression module is a two-layer fully-connected network of dimensions $1024$ and $48$, which is referred to as \textbf{\textit{Regression\#1}}. To further validate the flexibility and complementarity of our proposed framework to other approaches, we also try to use state-of-the-art 3D pose estimators~\cite{martinez2017simple,DBLP:conf/eccv/SunXWLW18} as the regression components. The learnt representation $G$, behaves as a 3D structure prior, is injected into their frameworks. These two configurations are referred to as \textbf{\textit{Regression\#2}} and \textbf{\textit{Regression\#3}} respectively. Note that, in order to evaluate the robustness and flexibility of the proposed geometry representation in a straightforward manner, we \textit{only} forward the geometry representation $G$ to fully connection layers to match the feature dimension of baselines, and then directly do element-wise sum with baselines, instead of designing sophisticated feature fusion mechanism to potentially better fuse the representation with original features. All the experiments are conducted on Titan X GPUs. Please refer to the supplemental materials for architecture details.

\begin{figure}[]
   \begin{minipage}{0.42\linewidth}
      \centerline{\includegraphics[width=1.15\linewidth]{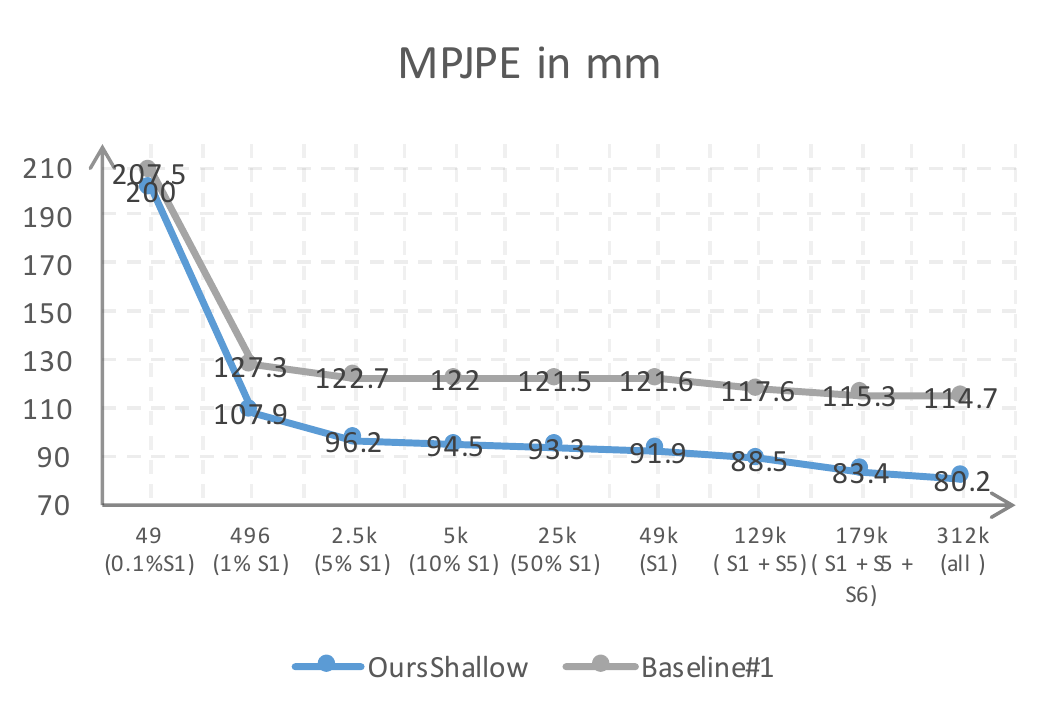}}
      \centerline{(a)}
   \end{minipage}
   \hfill
   \begin{minipage}{.42\linewidth}
      \centerline{\includegraphics[width=1.15\linewidth]{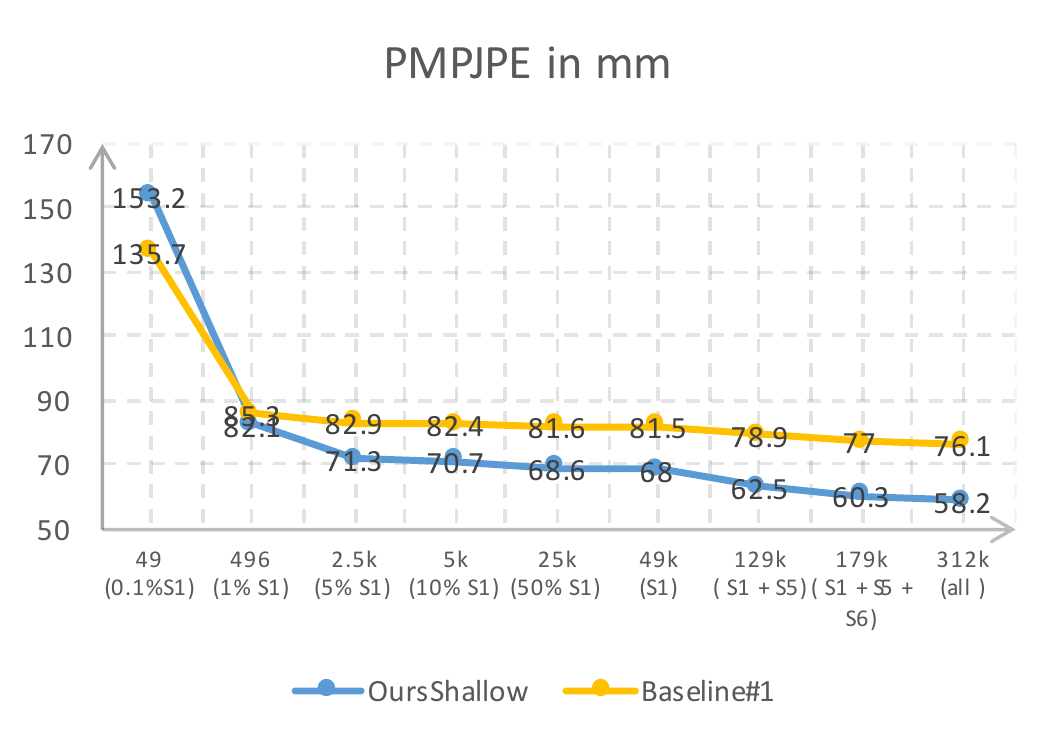}}
      \centerline{(b)}
   \end{minipage}
   \vspace {-0.2cm}
   \caption{\label{fig4}\small{Evaluation on the Human3.6M using different number of training data. (a) presents the results under MPJPE metric. (b) presents the results under PMPJPE metric.}}
   \vspace {-0.55cm}
\end{figure}

\textbf{Results on Human3.6M.}
We firstly validate the effectiveness of learnt representation $G$ to 3D human pose estimation task, on the condition of using different amount of 3D annotated samples (under \textit{Protocol\#1}) to train the regression module. We adopt \textit{Regression\#1} as the regressor with \textit{only} $G$ as the input. The configuration is referred as \textit{OursShallow}. Since only 2D annotation is utilized to learn $G$, we also list the performances of directly regressing 3D pose coordinates from 2D detections with the same regressor, which is referred to \textit{Baseline\#1}. Figure~\ref{fig4} shows the results. The  phenomenon is consistent on both MPJPE and PMPJPE metrics. Given only about $500$ annotated training samples, our method achieves $17.98\%$ relative improvements than \textit{Baseline\#1} on MPJPE, and $3.90\%$ on PMPJPE. The margin becomes larger when more annotated samples are used for training. Our general improvements over different setting demonstrate the robustness of the learnt representation to different amount of 3D training samples. We also perform above experiments on \textit{Regression\#2} and \textit{Regression\#3} to further verify the effectiveness of the learnt representation to strong baselines (For space saving, the detail results are shown in the supplementary material). Under fewer amount of training samples, our proposed representation could help improve the performance of baselines to comparable results with the one trained on a larger amount of samples by themselves.

We then evaluate the models under all three protocols to demonstrate the effectiveness and flexibility of learnt representation $G$ \textit{as a robust 3D prior} to different 3D human pose estimation methods. Table~\ref{tb1} reports the comparison with current state-of-the-arts. We draw two key observations as follows: $(1)$ Directly regressing 3D poses with \textit{only} learnt geometry representation $G$ as input and simple 2-layer fc architecture (Ours+ Regression\#1) could achieves reasonable 3D pose estimation results. $(2)$ As a 3D geometry prior, $G$ could easily help improving the performance of different backbones coherently, achieving state-of-the-art results under all three protocols. Even on the strong baseline like ~\cite{DBLP:conf/eccv/SunXWLW18}, which is the most state-of-the-art, the model (Ours+Regression\#3) could still have $7\%$ inprovements, achieving $46.3$ of mm of error.

\textbf{Ablation Study.}
We conduct ablation experiments on the Human3.6M dataset under \textit{Protocol\#1} to verify the effectiveness of different components of our method. The overall results are shown in Figure~\ref{fig5}. The notations and comparison are as follows:
\begin{itemize}
   \item{\textbf{BL}} refers to the 3D pose estimator without learnt representation $G$. We regard this model as the baseline model of our framework. We train the baseline with its public implementation~\cite{DBLP:conf/eccv/SunXWLW18}. The mean error of the baseline is $49.8mm$.

   \item{\textbf{BL+I\_SG}} refers to the use of \textit{raw} images to train the \textit{$generator(\cdot,\cdot)$}. We observe a drop of performance ($49.8mm\rightarrow52.6mm$), which is even worse than the baseline model. This result suggests that the raw image-based view synthesis mechanism could not facilitate the encoding of the representation due to the lack of the distilling step to distill unnecessary factors (\eg,appearance, lighting, and background).

   \item{\textbf{BL+AE}} refers to the configuration that the source and target domain are \textit{same} during the training of \textit{$generator(\cdot,\cdot)$}. The mean error is $49.9mm$, which is almost the same with the baseline. This result suggests that the latent codes of autoencoding could not provide more valid information than a pure 2D coordinate information, if there is no special mechanism incorporated in.
   \begin{figure}[t]
      \begin{center}
         \vspace {-0.45cm}
         \includegraphics[width=0.8\linewidth, angle=0]{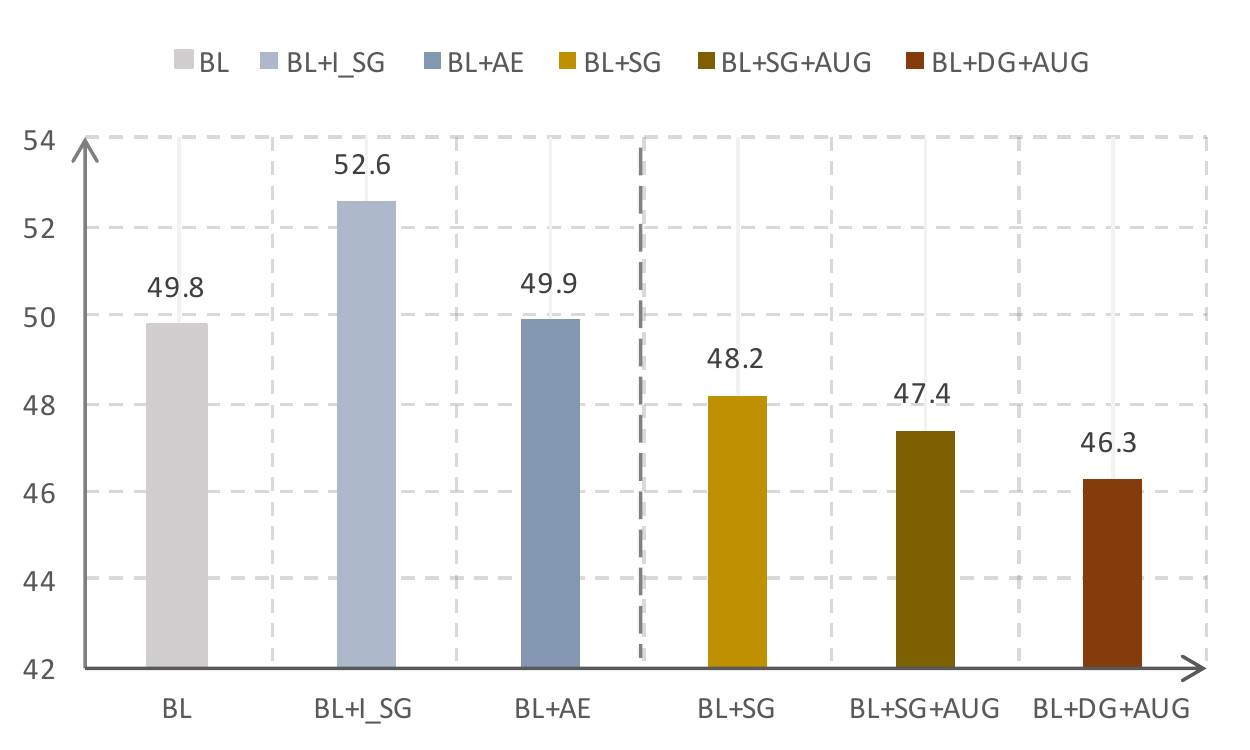}
      \end{center}
      \vspace{-0.5cm}
      \caption{\label{fig5}\small{Ablation studies on different components in our method. The evaluation is performed on \textit{Human3.6M} under \textit{Protocol\#1 }with MPJPE metric.}}
      \vskip -0.7cm
   \end{figure}

   \begin{figure*}[t!]
      \begin{center}
         \vspace {-0.5cm}
         \includegraphics[width=0.9\linewidth, angle=0]{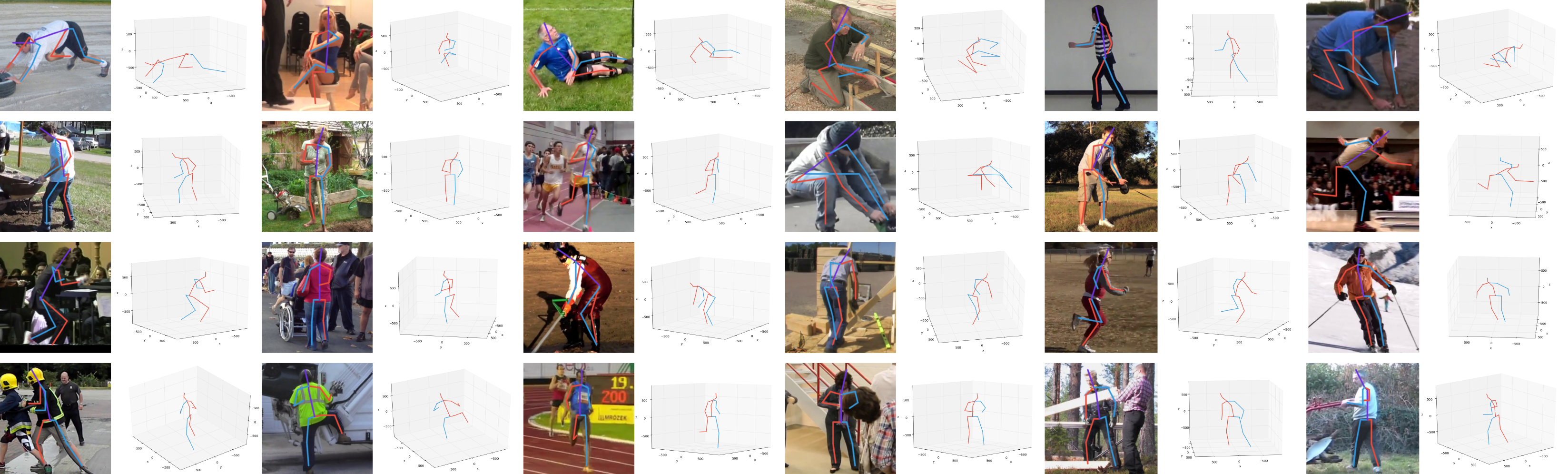}
      \end{center}
      \vskip -0.3cm
      \caption{\label{fig6}\small{Qualitative results of our approach on the test split of in-the-wild MPII human pose dataset. Best viewed in color.}}
      \vskip -0.5cm
   \end{figure*}

   \item{\textbf{BL+SG}} refers to the model that injecting learnt representation $G$ to the baseline network as a 3D structure prior, where $G$ is learnt \emph{without} representation consistency constraint. Simply adding the learnt $G$ to the baseline network by concatenation operation instead of any sophisticated fusion mechanism, the model reduces the error by $3.2\%(49.8mm\rightarrow48.2mm)$. This validates the effectiveness and flexibility of our framework to learn the geometry representation in the articulated human body. Moreover, comparing with the results on {BL+I\_SG}, {BL+SG} shows 2D skeleton maps could provide sufficient information to learn the geometry representation.

   \item{\textbf{BL+SG+AUG}} refers to the use of data augmentation by virtual cameras. The augmentation  provides $1.6\%$ lower mean error compared with ~`BL+SG'. In the ablation study that shown in supplemental materials, the augmentation on other baselines show similar results of relative improvements.
   
   \item{\textbf{BL+DG+AUG}} refers to the use of representation consistency constraint. We see a $2.3\%$ error drop ($47.4mm\rightarrow46.3mm$), showing that our proposed consistency constraint indeed increase the robustness of the geometry representation $G$. The constraints that conventionally designed in multi-view approaches, \eg epipolar divergence~\cite{DBLP:journals/corr/abs-1806-00104} and multi-view consistency~\cite{rhodin2018learning}, require iterative optimization-based method, like RANSAC, to initialize the process. In contrast, our representation consistency constraint is straightforward and purely feed-forward, which is easier to train and implement.
   
\end{itemize}

We further illustrate the ablation study on the configuration of \textit{Regression\#1} and \textit{Regression\#2}. The observation is similar to the results shown in Figure~\ref{fig5}, while the relative improvements among different components are more significant. Please refer to supplemental materials.

\textbf{Cross-Domain Generalization.}
Here, we perform three types of cross-dataset evaluation to further verify some merits of our approaches. 

We first demonstrate the generalization ability of the learnt representation between domains quantitatively. Table~\ref{tb3} reports the results of the configuration that training on Human3.6M and then testing on INF-3DHP. Following~\cite{mehta2017monocular,yang20183d}, we use AUC and PCK as the evaluation metrics. As can be seen from the results, our model with different regressors present consistent improvements to their baselines in most cases, which demonstrates the learnt geometry representation could improve the generalization ability of subsequent pose estimator significantly for its robust to new camera views and unseen poses.

\begin{table}[h]{}
   \begin{center}
      \resizebox{\columnwidth}{!}{%
         \begin{tabular}{c|cccccc|ccc}
            \Xhline{0.9pt}
            &\cite{mehta2017monocular}&\cite{zhou2017towards}&\cite{yang20183d}&R\#1&R\#2\cite{martinez2017simple}&R\#3\cite{DBLP:conf/eccv/SunXWLW18}&Ours + R\#1&Ours + R\#2&Ours + R\#3\\
            \Xhline{0.5pt}
            PCK &64.7&50.1&69.0&41.0&68.0&68.4&61.4&68.7&\textbf{75.9}\\
            \Xhline{0.5pt} 
            AUC&31.7&21.6&32.0&17.1&34.7&29.4&29.4&34.6&\textbf{36.3}\\
            \Xhline{0.9pt}
         \end{tabular}%
      }
   \end{center}
   \vspace {-0.3cm}
   \caption{\label{tb3}\small{Cross-dataset comparison with state-of-the-arts on the MPI-INF-3DHP dataset with PCK and AUC metrics. R\#* indicates Regression\#*.}}
   \vskip -0.4cm
\end{table} 


We then demonstrate the generalization ability of our model to the unconstrained environment qualitatively. Figure~\ref{fig6} shows the sampled results on the test split of MPII dataset, where the model is trained on Human3.6M dataset. As can be seen from the figure, our method is able to accurately predict 3D pose for in-the-wild images.

\begin{figure}[h]
   \begin{center}
      \vspace {-0.2cm}
      \includegraphics[width=1\linewidth, angle=0]{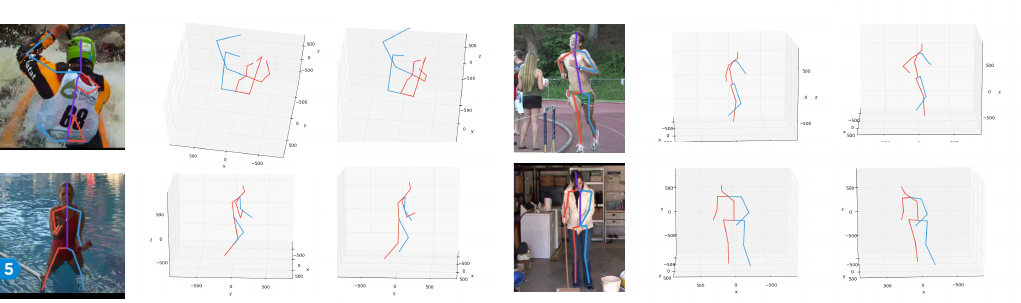}
   \end{center}
   \vspace {-0.6cm}
   \caption{\label{fig7}\small{Qualitative comparison on the MPII dataset. The second column shows the predictions of training on the Human3.6M dataset. The third column shows the predictions of cross-dataset training.}}
   \vskip -0.7cm
\end{figure}

Finally, we present the benefit of eliminating the inter-dataset variation to 3D human pose estimation. Since our framework breaks the gap of inter-dataset variation, different 3D human pose benchmarks could be trained together to increase the diversity. As shown in Figure~\ref{fig7}, cross-dataset training (Human3.6M + MPI-INF-3DHP) shows better robustness than single-dataset training (Human3.6M) on some unseen poses of the MPII dataset.

\section{Conclusion}
We have presented a weakly-supervised method of learning a geometry-aware representation for 3D human pose estimation. Our method is novel in that we take a radically different approach to learn the geometry representation under multi-view setting. Specifically, we leverage view synthesis to distill shared representation in the latent space with only the usage of 2D annotation and simple representation consistency constraint, which provides a new aspect to learn the representation with fewer annotation efforts and simpler network architecture. Meanwhile, we bridge different 3D human pose datasets by introducing a skeleton-based encoder-decoder. Experimental results validate the effectiveness and flexibility of the proposed framework on 3D human pose estimation task.

{\small
\bibliographystyle{ieee_fullname}
\bibliography{egbib}
}
\end{document}